\documentclass[conference]{IEEEtran}
\IEEEoverridecommandlockouts
\usepackage[colorinlistoftodos]{todonotes}
\usepackage{cite}
\usepackage{amsmath,amssymb,amsfonts}
\usepackage{algorithmic}
\usepackage{graphicx}
\usepackage{textcomp}
\usepackage{float}
\usepackage{xcolor}
\def\BibTeX{{\rm B\kern-.05em{\sc i\kern-.025em b}\kern-.08em
    T\kern-.1667em\lower.7ex\hbox{E}\kern-.125emX}}
\begin{document}

\title{DeepMorph: A System for Hiding Bitstrings in Morphable Vector Drawings\\
\thanks{This project was financially supported by the Danish Ministry of Higher Education and Science.}
}

\author{\IEEEauthorblockN{
Søren Rasmussen\IEEEauthorrefmark{1},
Karsten Østergaard Noe\IEEEauthorrefmark{1},
Oliver Gyldenberg Hjermitslev\IEEEauthorrefmark{1},
Henrik Pedersen\IEEEauthorrefmark{1},
}
\IEEEauthorblockA{
\IEEEauthorrefmark{1}Visual Computing Lab, Alexandra Institute, Aarhus, Denmark\\
Email: henrik.pedersen@alexandra.dk}}

\maketitle

\begin{abstract}
We introduce DeepMorph, an information embedding technique for vector drawings.
Provided a vector drawing, such as a Scalable Vector Graphics (SVG) file, our method embeds bitstrings in the image by perturbing the drawing primitives (lines, circles, etc.).
This results in a morphed image that can be decoded to recover the original bitstring.
The use-case is similar to that of the well-known QR code, but our solution provides creatives with artistic freedom to transfer digital information via drawings of their own design.
The method comprises two neural networks, which are trained jointly: an encoder network that transforms a bitstring into a perturbation of the drawing primitives, and a decoder network that recovers the bitstring from an image of the morphed drawing.
To enable end-to-end training via back propagation, we introduce a soft rasterizer, which is differentiable with respect to perturbations of the drawing primitives.
In order to add robustness towards real-world image capture conditions, image corruptions are injected between the soft rasterizer and the decoder.
Further, the addition of an object detection and camera pose estimation system enables decoding of drawings in complex scenes as well as use of the drawings as markers for use in augmented reality applications.
We demonstrate that our method reliably recovers bitstrings from real-world photos of printed drawings, thereby providing a novel solution for creatives to transfer digital information via artistic imagery.
\end{abstract}


\section{Introduction}
In recent years, modern society has relied on barcodes and QR codes for a wide variety of data-transfer tasks, such as redirecting users to web pages and initiating mobile contactless payments.
These methods, however, are immutable and visually unappealing, and they clearly reveal that data exists and how to decode it. Image steganography attempts to solve these issues by embedding secret data into seemingly ordinary image files, but struggles with preserving the embedded data during physical transmission (i.e., printing or display and subsequent image capture using a camera).

Our goal is to devise a method that seamlessly embeds messages in images in such a way that the message can be decoded from a captured photo of the embedded image. The use-case is similar to that of the QR code and similar technologies. Our approach targets creatives, such as artists, designers, and illustrators, who wish to create aesthetically pleasing alternatives to barcodes, enabling them to transfer digital information via artistic imagery.

\begin{figure}[t]
\centerline{\includegraphics[width=\columnwidth]{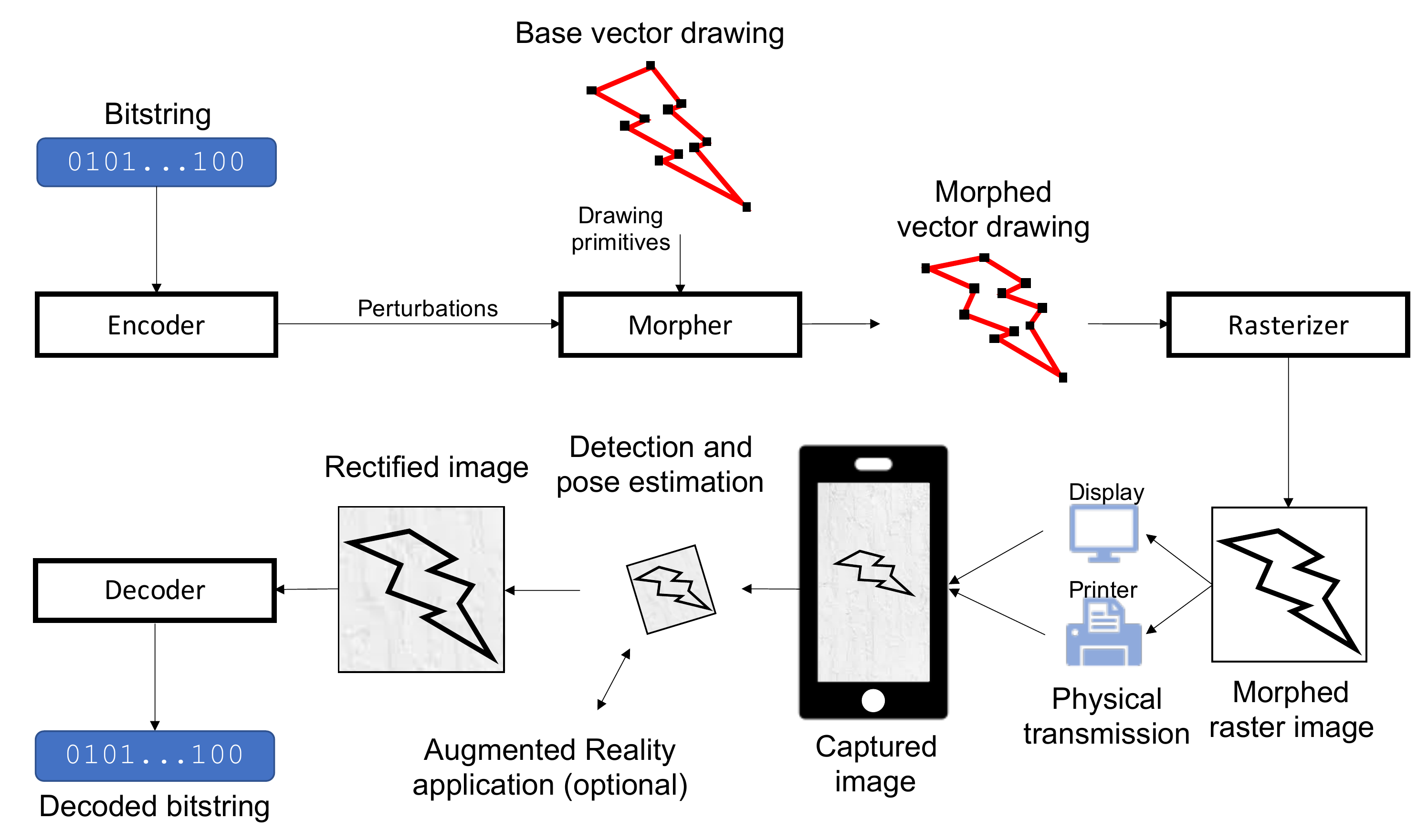}}
\caption{Overview of the proposed system for encoding and decoding bitstrings in vector drawings.}
\label{fig1}
\end{figure}
To provide maximal flexibility and artistic freedom, our method is based on vector graphics. Vector graphics use drawing primitives such as points, lines, curves, and polygons - all of which are based on mathematical expressions — to represent images. The primitives are made from vectors (also called paths or strokes) that lead through control points with a definite position on the x and y axes of the work plane. In addition, each path may be assigned a stroke color, shape, thickness, and fill. These primitives describe solely how the vector graphics should be drawn; a process known as rasterization. The core idea of our approach is to hide the bitstring by perturbing the drawing primitives. This results in a morphed raster image, which can be decoded to recover the original bitstring. To preserve the overall layout of the vector drawing the perturbations are constrained to be within bounds specified by the artist.

Figure 1 presents an overview of our system, which we call DeepMorph, in a typical use context. The input is a bitstring, which represents the data to be transferred (e.g., a hyperlink or a product ID associated with a unique bitstring). The bitstring is used to morph a base vector drawing, created in advance by the provider of the system. The morphing is determined by a neural network encoder that takes as input the bitstring and outputs perturbations of the drawing primitives. The morphed vector drawing is then rasterized to produce a morphed raster image that can be printed or displayed on a screen (physical transmission). When the end-user captures a photo of the physical image, a neural network decoder retrieves the unique bitstring, which is then used to follow e.g. a hyperlink or view product information. Our method relies on object detection and camera pose estimation to correctly decode the bitstring, and therefore, as an extra benefit, the vector drawing also functions as a marker for augmented reality applications.

\begin{figure}[t]
\centerline{\includegraphics[width=.7\columnwidth]{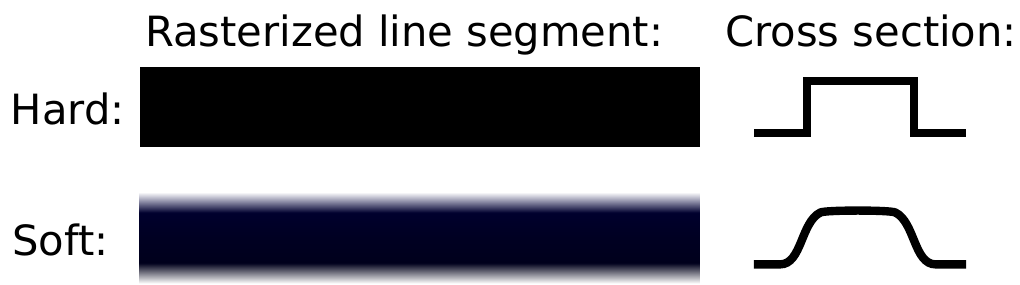}}
\caption{
    Illustration of hard vs. soft rasterization of a simple line segment.
    Contrary to the hard rasterization, the pixels of the soft rasterization are differentiable with respect to the vector definition of the line (i.e. the endpoint coordinates)
}
\label{fig_softraster}
\end{figure}

Data transmission via images has a long history in the fields of image steganography, watermarking, and barcode design. Here, we present the first deep neural network architecture for this problem using vector graphics. To enable end-to-end training via backpropagation, our solution draws on theory from neural rendering to implement a soft rasterizer that is differentiable with respect to the primitives of the base vector drawing (see Fig.~\ref{fig_softraster}). Furthermore, we inject pixelwise and spatial corruptions between the encoder and decoder to simulate the distortions resulting from the image capture. The result is robust retrieval of 24 encoded bits in real-world conditions.

\section{Related Work}
\subsection{Steganography}
Image steganography is the practice of hiding secret data by embedding it into an image. Typical applications include transmission of sensitive data and watermarking for copyright protection. Many methods for image steganograpy in digital images assume that the image is perfectly digitally transferred. This makes it possible to hide information in the digital file - for instance by modifying the least significant bits of pixel colors or introducing variations in color and luminosity (e.g., in jpeg compression). The reader is referred to \cite{Cheddad2010} for an extensive survey on methods for digital image steganography.

In recent years the use of deep neural networks has been dominating the scientific field of image steganography.
An overview can be found in the survey paper by Meng et al. on methods for storing secret information in images using deep learning \cite{cmes.2018.04765}. Some authors have focused on the use of deep learning for improving pure digital image steganography \cite{yang2018stegnet, Duan2019ReversibleIS}. However in the use-case of our approach, the image undergoes an analog transmission because it is printed or displayed on a screen from which a photograph is taken with e.g. a mobile device. Therefore, we need to ensure that the steganographic information is preserved in the display and re-digitization process. Multiple authors have presented results that work towards this goal. Baluja \cite{baluja2017deepstega} present a deep learning based approach for image steganography that relaxes the requirement for perfect digital information transfer. 
Zhu et al. \cite{Zhu2018HiDDeNHD} presents a deep learning approach achieving robustness towards Gaussian blurring, cropping and JPEG compression. Full support for printing and re-digitizing has been demonstrated by Tancik et al \cite{Tancik2019StegaStampIH} who utilize deep learning for encoding and decoding of arbitrary bitstrings into natural images. In our work we have based our approach on vector drawings instead, because this allows explicit artistic control over visual appearance and variation.

A few authors have approached the task of storing steganograpic information directly in digital vector graphics files \cite{mados2014svgstego,almutairi2019svglsbstego}. However to support our use-case we cannot rely on permutations of the digital file for information transfer but must instead provide robustness towards physical transmission, such as image printing and re-digitization.
This robustness was achieved by Xiao et al. who encoded information by creating a codebook of different glyphs representing pertubations of vector letters in a font \cite{Xiao2017fontcode}. They used deep learning for decoding the hidden information and also for selecting the glyphs from a font manifold to include in the codebook. In contrast we also use deep learning for training the optimal pertubation of vector geometry in order to encode hidden information. To the best of our knowledge, our work is the first approach directly aimed at combining vector geometry pertubation and deep learning for storing steganographic information in vector drawings in this manner. 

The task of disclosing the hidden information is called \textit{steganalysis} and a significant focus of much work \cite{baluja2017deepstega, Zhang2018deepstega} is to evade the unwanted disclosure of the hidden information. Making steganalysis impossible is not the main focus of our approach, although the deep learning based encoding scheme is very hard to reverse engineer without access to the encoder network.

\subsection{Watermarking}
Watermarking is a subfield of steganography that is used to hide proprietary information in digital media like images, music, and video. The main requirements of digital image watermarking are transparency, robustness, and capacity \cite{alattar2000smart, Cox}, all of which are important in our use-case as well.

Transparency means that the watermarked image should be perceptually similar to the original. This is often solved by embedding the watermark in the frequency domain representation of the image \cite{Cox, Potdar}. In our case, the creator/artist explicitly specifies the degree of transparency by putting bounds on the perturbations that the drawing primitives are allowed to undergo.

Robustness refers to the ability to detect the watermark after common signal processing operations (e.g., contrast enhancement and lossy compression), geometric distortions (e.g., scaling and rotation), unauthorized attacks, and printing. Rather than hand-crafting a solution to increase robustness, our method automatically learns to ignore the types of corruptions that we deliberately inject between the encoder and the decoder. These include both pixelwise and spatial corruptions. However, we explicitly handle large distortions in image perspective by estimating the pose of the camera relative to the image (see section ~\ref{sec:pose}).

Finally, capacity means that the watermark should be able to carry enough information to represent the uniqueness of the image. In our case the capacity is largely determined by the number of drawing primitives that can be perturbed, as well as their user-defined bounds (e.g., if control points are allowed to move more the capacity generally increases).

\subsection{Fiducial Markers and Barcodes}

The traditional one dimensional bar code consisting of stripes of varying thickness is popular for transmitting a small message because it requires only very simple laser hardware to decode. With the surge of the smartphone, a number of 2D square based fiducial markers like the QR code have been used for message transfer. Some of these markers, like ARTags or ArUco codes \cite{GARRIDOJURADOfiducialMarkers}, have specifically been designed for defining a reference coordinate system (6D pose) for use in an augmented reality (AR) context. 

Besides information transfer, an additional benefit of our system is that it can replace these conventional markers for use in an AR context while allowing a higher artistic freedom on the visual appearance. We achieve this by employing deep learning for object detection and pose estimation as explained in section \ref{sec:pose}.

\subsection{Neural Rendering}
Neural rendering exists in the overlap between computer graphics and deep learning \cite{tewari2020state}. The technique employs differentiable computer graphics modules such that networks can learn to generate graphical representations based on various input. Neural rendering has been explored as a solution to a multitude of problems, including novel view synthesis \cite{sitzmann2018deepvoxels, sitzmann2019scene, xu2019deep}, image manipulation \cite{bau2019semantic, brock2016neural}, and facial reenactment \cite{thies2016face, zakharov2019fewshot}, among others. Key for neural rendering is that traditional computer graphics operations are replaced with differentiable approximations, e.g., replacing a step function with a sigmoid. This issue comes through during rasterization, the process in which vector-based primitives are mapped to pixel grids, as rasterization is a discrete operation.
The soft rasterizer used in our system is partly inspired by the neural mesh renderer \cite{kato2018neural}, where gradients flow using edge blurring during backpropagation, allowing the system to re-render learned geometric primitive representations.

\subsection{Object Detection and Pose Estimation}
\label{sec:pose}

To correct for the spatial distortions that result from variation in camera position and rotation, our system detects and rectifies the encoded image before decoding it. Traditional computer vision rectifies images by locating pre-defined keypoints in the image (e.g., using the Scale-Invariant Feature Transform (SIFT) \cite{lowe2004distinctive}) and subsequently estimating and applying a suitable perspective transform. Further, if the intrinsic parameters of the camera are known in advance, and if the physical 3D position of each keypoint on the object being imaged is also known, the rotation and translation (i.e., extrinsic parameters) of the camera can also be determined \cite{pnp}, thereby enabling the use of the encoded image as a marker for AR applications.

Although traditional techniques like SIFT are designed to be invariant to a wide range of image transformations (like changes in viewpoint and brightness), they ultimately fail in our use-case because the vector drawing is allowed to morph. Modern approaches based on convolutional neural networks (CNNs) overcome this problem by learning semantic representations of the objects being imaged, enabling detection of semantic keypoints under much more varying viewing conditions \cite{newell2016stacked, wei2016convolutional}. The type of CNN being used is typically an image-to-image network that outputs heatmaps that help detect the keypoints. Because the computations of this type of network can be relatively expensive, it is commonplace to first apply an object detection network that identifies a bounding box around the object of interest such that the image-to-image network can be applied on a smaller cropout to speed up computation time \cite{he2017mask}. Our system uses a fast single-shot object detector tailored for mobile devices \cite{liu2016ssd}.

\section{Methods}
\subsection{Morphable Vector Drawings}
SVG drawings consist of nested groups of parametrically defined drawing primitives, such as lines and circles, and parametrically defined transformations of these groups.
For example, a simple straight line may be defined by its endpoint coordinates, thickness and color. For any given drawing, we refer to the combined set of primitive- and group-parameters simply as the drawing parameters. For most drawings, slight changes in the parameters do not significantly alter the artistic expression and are often barely noticeable. We exploit this fact to encode a bitstring in a given drawing by perturbing the parameters within predefined bounds specified on a per-primitive (affine transformations and per-point translation) and per-group basis (affine transformations). For example, the artist may specify that the endpoints of a line may move within a radius of 2 units around their nominal position, or that a group may be scaled within a certain range. The result is a drawing with some number of variables, $N_v$, defining the parameter perturbations, each subject to predefined bounds. Currently, the system supports a small subset of the SVG primitives, namely lines, paths (linear and arch segments), circles, half-circle and polygons. Also, the system is restricted to producing black-and-white drawings.
\begin{figure}[t]
\centerline{\includegraphics[width=\columnwidth]{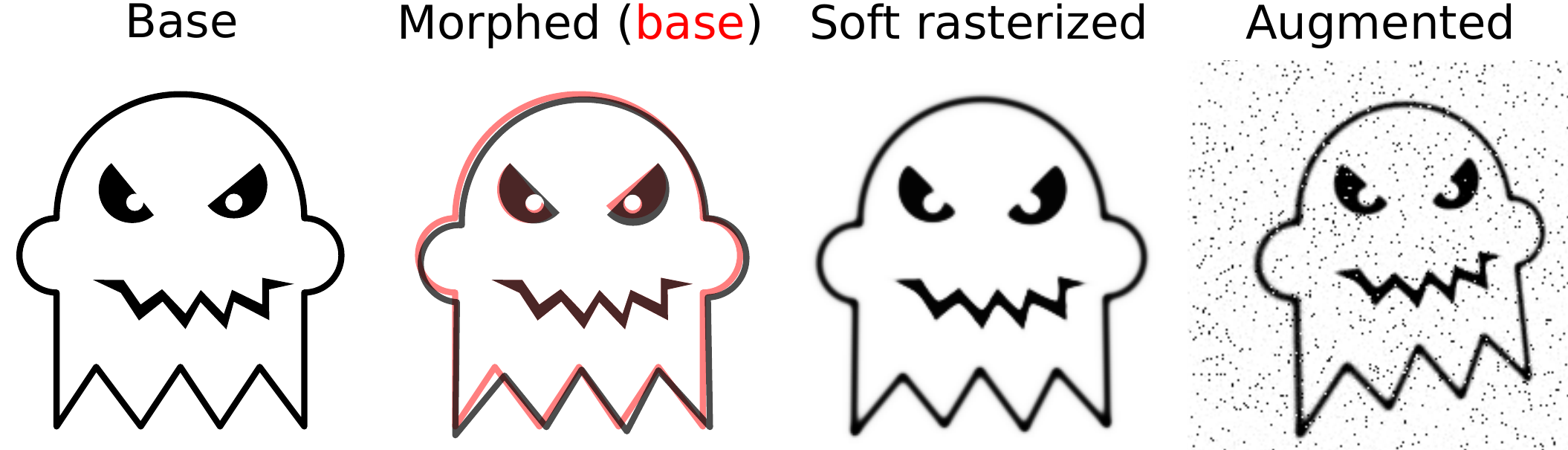}}
\caption{Example showing morphing of vector drawings and training data generation for a 24-bit code.
         From the left:
            1\textsuperscript{st}: Base vector drawing.
            2\textsuperscript{nd}: Morphed vector drawing, overlaid on the base drawing (red) for comparison.
            3\textsuperscript{rd}: Soft rasterization of the morphed vector drawing.
            4\textsuperscript{th}: Augmentation applied to soft rasterization.
}
\label{fig_ghosts}
\end{figure}

\subsection{Encoder}
In order to encode a bitstring $b$ of length $N_b$ in a drawing with $N_v$ variables, the bitstring is passed through a simple neural network encoder. The encoder consists of a single, trainable, linear layer $A$ of shape $(N_v \times N_b)$ followed by a sigmoid activation, $\sigma$, and a per-variable transformation, $T$ which maps the range of the sigmoid activation to the bounds of the given variable. The result is then combined with the original drawing parameters $p$, to yield the perturbed parameters $\hat{p}$:
\begin{equation}
    \hat{p} = \operatorname{perturb}(p,T[\sigma(Ab)])
\end{equation}
where $\operatorname{perturb}$ is an affine function, defined at group-level (affine transformations) or at primitive-level (affine transformations and per-point translation). A new drawing can now be generated by simply swapping $\hat{p}$ for $p$ in the vector drawing definition.

\subsection{Decoder}
Extraction of a bitstring from a drawing is a multi-label binary $N_b$-class classification problem. Importantly, the input to the decoder is not the vector drawing itself, but a morphed raster image that has been rendered from the perturbed parameters $\hat{p}$. In the following, we denote the rasterized image $I_b^A$ to indicate that it was formed from bitstring $b$ using encoder weights $A$. The decoder is implemented as a small CNN classifier consisting of a few convolutional layers, followed by one or more fully connected layers. The output is an $N_b$-dimensional vector
\begin{equation}
    \hat{x} = \sigma(\operatorname{CNN}[I_b^A])
\end{equation}
Due to the sigmoid activation the $i$'th element of $\hat{x}$ can be interpreted as the pseudo-probability that the $i$'th element of $b$ is 1. Hence, the recovered bitstring is obtained by thresholding $\hat{x}$:
\begin{equation}
\hat{b}_i =
  \begin{cases}
    0   & \quad \text{if } \hat{x}_i \leq 0.5\\
    1   & \quad \text{if } \hat{x}_i > 0.5
  \end{cases}
\end{equation}
where subscript $i$ refers to the $i$'th element of the corresponding vector.

Although modern CNNs typically contain in the order to 10-100 convolutional layers, our experience is that a small three-layer CNN with a two-layer fully connected classification head is both effective and relatively computationally light-weight.

\subsection{Enabling End-to-End Training}
In principle, the decoder can be trained to extract bitstrings from drawings produced by a randomly initialized encoder. In practice, however, this yields poor results. To enable end-to-end training, and in order to learn a robust encoder/decoder combination, we introduce a differentiable "soft" rasterization engine to link the encoder output, $\hat{p}$ with the decoder input $I$. This rasterization engine is capable of rendering the 2D vector graphics in a form, which closely resembles the "hard" output of a conventional nondifferentiable vector graphics rasterizer, while at the same time allowing gradient flow between the encoder and decoder.

The rasterization process is as follows:
Each drawing primitive has an associated differentiable signed distance function (SDF), which is positive in the interior of the primitive, zero on the border and negative outside.
By computing each SDF over a pixel grid and applying a sigmoid function with a sharpening parameter $s$, each primitive is rendered in a differentiable manner, yielding $L$ rasterizations, $\left\{r_0, r_1, \dots, r_{L-1}\right\}$.
To form the final image, $I$, the rasterized primitives are combined on a zero-valued (white) canvas in an additive fashion, with black- and white-colored primitives represented by positive and negative values, respectively, and the result is squashed to the range $[0;1]$:
\begin{equation}
    I = \operatorname{squash}_{[0;1]}\left(
                                \sum_{l=0}^{L-1}c\cdot\sigma(sr_i)
                           \right)
\end{equation}
where
\begin{equation*}
    c = \begin{cases}
        -1 & \quad\text{for white-colored primitives}\\
        1  & \quad\text{for black-colored primitives}
    \end{cases}
\end{equation*}
and
\begin{equation}
    \operatorname{squash}_{[0;1]}(x) = \sigma[t(x-0.5)]
\end{equation}
with $t$ being a sharpening parameter.

This approach has the obvious shortcoming that layering is not supported. 
For example, adding a white line on top of two overlapping black lines yields a black line, i.e.
\begin{equation*}
    B+B+W=1+1-1=1 \text{ (black)}
\end{equation*}
where $B=1$ and $W=-1$ represent the black and white features, respectively.
To mitigate this, bottom-up layering is enabled by the use of intermediate squashing functions, i.e.:
\begin{equation*}
    \operatorname{squash}_{[0;1]}(B+B) + W = B+W = 0 \text{ (white)}
\end{equation*}

\subsection{Training}
Empirically we observe that training the full network from scratch is slow and prone to divergence. This is solved by applying a two-stage training schedule. In stage 1, the encoder weights are locked at their randomly initialized state, and the decoder is trained until performance plateaus. Then, in stage 2, the encoder weights are unlocked, and the system is trained end-to-end.
The loss function is a sigmoid cross-entropy, as commonly used in multi-label classification.

To add robustness to the various artifacts, introduced by real-world cameras, we add random augmentation to the images between the rasterizer and the decoder. Note that in contrast to a more conventional setting, where the augmentation is applied before the input stage of the network, this setting requires the augmentations to be differentiable with respect to the image.
Specifically, we apply spatial transforms (perspective, scaling, rotation, translation) and pixel intensity noise (scaling, offset, gaussian noise and salt-and-pepper noise).

\subsection{Detector}
For real world use we must detect and rectify the encoded image before decoding it. To this end, we employ a lightweight CNN-based object detection model trained on randomly generated renderings of the vector drawing. The randomness consists of perturbing the drawing with random bitstrings, and adding artificial background images and noise. The object detector is a SSD network implemented in Tensorflow and trained using an established training scheme \cite{tensorflow_2020}.

The result of the object detection stage is a cutout of the encoded image with some perspective warp. To rectify the image, a number of imperturbable keypoints in the drawing are selected. These keypoints are detected using a second CNN similar to \cite{newell2016stacked} which generates a heatmap for each keypoint, followed by a soft-argmax operation to compute the keypoint positions with sub-pixel accuracy \cite{sun2018integral}. We use an iterative PnP solver \cite{pnp} with respect to the known imperturbable markers to retrieve the camera pose.

With the cutout aligned to closely resemble its original vector graphic orientation, we perform a series of preprocessing operations to prepare it for decoding. First, the cutout is blurred using a standard Gauss kernel of size 3x3 pixels and sigma 0.95. This blurred image is thresholded using Otsu's thresholding \cite{otsu1979threshold}, creating a binary representation of the graphic. A predefined mask, which coarsely surrounds the graphic, is used to mask the pixel values outside the figure as well. This sharply edged binary representation is yet again blurred using the same Gaussian kernel described earlier before being fed to the decoder network.


\section{Results and Discussion}
\begin{figure*}[hpt!]
\centerline{\includegraphics[width=\textwidth]{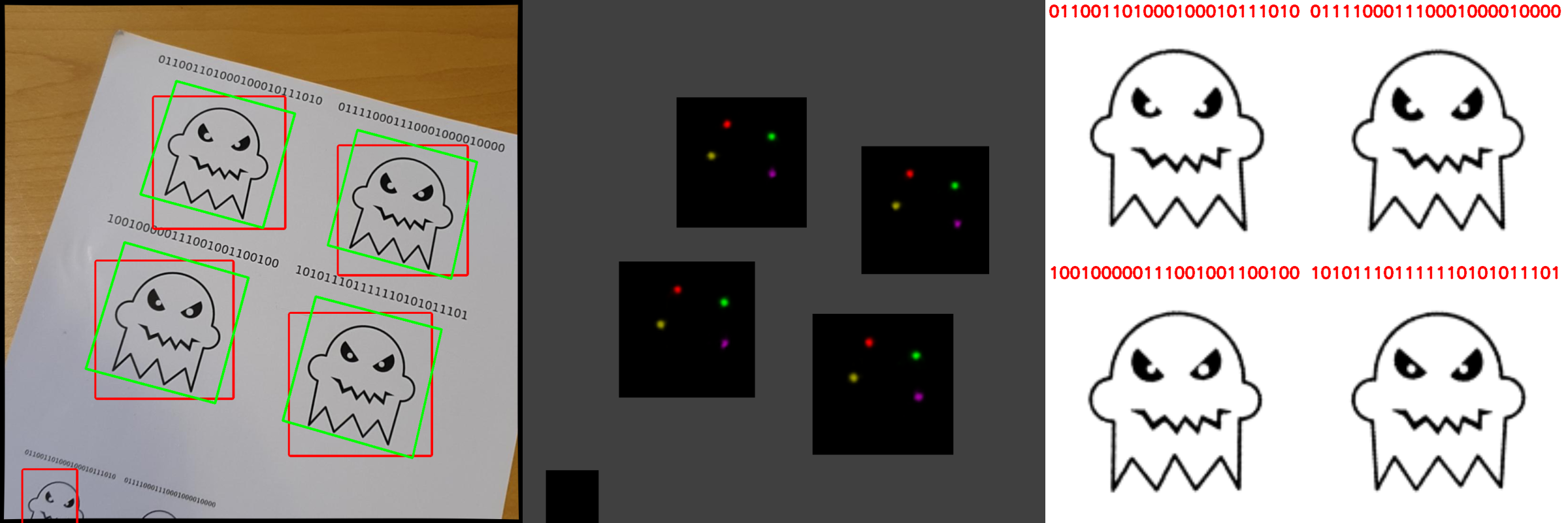}}
\caption{
    Real-world usage example of the drawing shown in Fig.~\ref{fig_ghosts}.
    Left: Printed input image. Bounding boxes, returned by the object detector, are shown in red. The green boxes show the marker pose, derived from the detected marker keypoints.
    Middle: Keypoint heatmaps for each detected marker (corners of eyes and mouth).
    Right: Rectified and pre-processed detections and decoded bitstrings.
    Note that the printed bitstrings are only for reference, and are not seen by the decoder.
}
\label{fig_realuse}
\end{figure*}

Fig.~\ref{fig_realuse} shows a real-world usage example of the full detection pipeline, including the object detector, pose estimation and the decoder.
Empirically, the system is robust towards varying perspectives, lighting conditions, dirt, etc., and provides reliable results for the decoded bitstrings.
Extensive testing in varying conditions is necessary to provide meaningful quantitative performance measures. Thus far, we have not observed that the method fails.
Approaches to further ensure reliability of the decoder include averaging predictions over time (at the cost of compute resources) and the use of checksums or error correcting codes, such as Hamming codes \cite{hamming1950error} (at the cost of reducing the number of information carrying bits per drawing). Additionally, using spatial transformer networks \cite{jaderberg2015spatial} would make image rectification an integral part of the neural network architecture. This will remove the need for fixing control points in the vector drawing for this operation, freeing up these control points to increase capacity for number of encoded bits. Further, the spatial transformer can be trained end-to-end with the encoder/decoder to increase robustness against perspective changes that are introduced during image capture.

A significant bottleneck in training the encoder-decoder networks end-to-end is the memory, required for gradient computations. Two especially important factors here are the drawing complexity and the raster resolution. This can be partially mitigated by implementation details such as pre-computating the rasterization of nonmorphable primitives (e.g. the mouth in  Fig.~\ref{fig_ghosts}) and by careful implementation of the rasterization functions, which remains a topic for further work.

\section{Conclusion}
We have presented DeepMorph, an end-to-end deep learning framework for encoding bitstrings in vector drawings, and a potential replacement for existing barcodes and QR codes, providing creatives with artistic freedom to transfer digital information via drawings of their own design. Joint training of the encoder and decoder is enabled through a differentiable "soft" rasterizer, and image augmentation is used during training to allow the decoder to generalize to real-world. Furthermore, the output of the required image rectification step can be used to estimate the camera pose, thereby allowing the vector drawing to used as a marker for AR applications. We demonstrate empirically that robust decoding of 24 bits, simultaneously with performing image rectification, is possible on printed images. Further experiments are required to determine the limitations of the framework, to guide creatives on how to make vector drawings with both high capacity and robust data transfer.

\bibliographystyle{plain}

\end{document}